# SFFDD: Deep Neural Network with Enriched Features for Failure Prediction with Its Application to Computer Disk Driver

Lanfa (Frank) Wang and Danjue Li

*Abstract*— A classification technique incorporating a novel feature derivation method is proposed for predicting failure of a system or device with multivariate time series sensor data. We treat the multivariate time series sensor data as images for both visualization and computation. Failure follows various patterns which are closely related to the root causes. Different pre-defined transformations are applied on the original sensors data to better characterize the failure patterns. In addition to feature derivation, ensemble method is used to further improve the performance. In addition, a general algorithm architecture of deep neural network is proposed to handle multiple types of data with less manual feature engineering. We apply the proposed method on the early predict failure of computer disk drive in order to improve storage systems availability and avoid data loss. The classification accuracy is largely improved with the enriched features, named smart features.

## 1. INTRODUCTION

Despite major efforts, high reliability remains a major challenge in running large-scale IT systems, and disaster prevention and cost of actual disasters make up a large fraction of the total cost of ownership. Computer disks are among the most frequently failing components in today's IT environments. Despite a set of defense mechanisms such as RAID, the availability and reliability of the system are still often impacted severely. Failure prediction in such circumstances would allow for the reduction of possible data loss and downtime costs through anticipated disk replacements.

This study focuses on researching on device failure prediction including, but not limited to, computer disk. Instead, we aim to explore potential methods of the similar problem for general device and system (if not for all devices and systems). Failure prediction is super important for critical devices such as flight engines and medical devices. Losses can be life, time, important data, expensive cost of device repair and so on.

The causes of failure include defects in design, process and quality during manufacturing, usage which is the underlying cause of a failure or that initiate a process that leads to failure. The types of failures due to usage include ageing, misuse, or mishandling. There are multiple patterns in general which leads to failure.

There are a few challenges in failure detection. Most critical devices have low failure rate. If the device fails frequently, it is easy to collect enough failure events to build accurate model.

For low failure rate devices, large dataset is required to build a reliable model because a minimum number of failure events is essential to capture the statics meaning and patterns. It typically requires collecting large dataset from a large number of devices over a long period of time. In most cases, the dataset is imbalances due to the nature of problem.

In general, a device follows certain patterns before it fails. These patterns make prediction and detection possible. However, one device can fail for various reasons and causes, therefore, there are many paths (failure patterns) in general. The more complicated the device, the more potential failure patterns while normal behaviors are similar. This character largely relaxes the failure detection task in general if the types of failure (or failure cause) is not a concern.

The third challenge is turn-on failure. The device may fail after it is turned on for short period of time. There is not enough data available for such an incident before it occurs. This type of failure closely relates to the device defect from manufacture and is excluded from the predictive model for operation.

Many of the techniques proposed so far mainly focused on feature selection and algorithm optimization. In this work, we present a novel method to derive new features specifically for failure predict to better represent the failure patterns. The performance can be largely improved with such enriched data.

Manual feature engineering is very time consuming for most machine learning tasks and it varies largely with the data. In this article, we propose a novel failure detection architecture with pre-defined automated feature engineering for general case aiming to reuse the algorithm as much as possible with minimized manual feature engineer work. Meanwhile, it can handle various types of data, including multivariate times series data, video, text and voice. Under this architecture, we name the algorithm **SFFDD** (**S**mart **F**ault and **F**ailure **D**eep **D**etector based on deep neural network). The proposed method also works with traditional machine learning algorithms such as Random Forest (RF) and Support Vector Machine (SVM). However, they can't handle multiple types of dataset simultaneously as neural network. In addition, neural network provides feature automatic extraction.

Here is a summary of our main contributions:

- A novel feature derivation method specially for failure prediction is proposed. The derived multiple smart features can be used to learn the failure in a deep way for various types of patterns. This ensures a robust single model for all brands because it can differentiate the normal and failed device in a deep way to catch the general pattern of all brands and models. This also offers one great advantage to leverage the available dataset when they are limited at early stage of the operation.

- Ensemble method further reduces the variance and improves the accuracy.
- Our methods also should work with other device failure problems with multivariate times series data, for instance IIOT sensors data.
- The proposed smart features can be directly used by various types of machine learning algorithms. Here we proposed an architecture of SFFDD based on CNN because it's generalized architecture and can take various types of dataset (numeric, text, video and voice) together into single analysis model for better prediction. Furthermore, the feature extraction layers automated extract features without manual feature engineering.

The rest of paper is organized as follows: Section 2 describes the disk driver dataset used for failure prediction. The overall architecture of **SFFDD** is introduced in Section 3. Section 4 describes smart features that are the core idea of **SFFDD**. Section 5 shows how to build a classifier step by step using smart features. Section 6 explores ensemble method for further improvement. Section 7 shows how to predict failure for different time window and Section 8 briefly summarizes related works for comparisons.

## 2. RELATED WORK

Traditional machine learning classifiers can be used for device failure prediction. In [1], Various algorithms Naïve Bayes, Support Vector Machine, Clustering are used for disk failure prediction with a best performance of 50.6% detection and 0% false alarm.

In [2], the authors came up with disk replacement prediction algorithm using S.M.A.R.T metrics with significant changepoint in time series and applied exponential moving average on the data for directly identifying the state of a device: healthy or faulty. Regularized Greedy Forest classifier (RGF), Gradient Boosted Decision Trees (GBDT), RF and SVM are used to build the model. The performance is sensitive to disk model. RGF achieves an accuracy of up to 92%, predicting failures with a time in advance of 10-15 days for particular disk model ST4000DM000 but drops largely for other models. Meanwhile, the prediction performance drops quickly with time in advance. A transfer learning technique is also explored enabling the creation of classifiers for HDD models with few data, providing an enhancement of up to 50% in accuracy, if compared to a classifier built with only the data at hand.

Aussel et al., [3] used the same dataset to perform hard drive failure prediction with SVM, RF and GBT. The dataset for all disk models is used and only nine raw S.M.A.R.T metrics are used to build the model: 5, 12, 187, 188, 189, 190, 198, 199 and 200. The precision and recall values based on the threshold of device failure within 10 days were recorded as almost 0.93 and 0.6 using cross validation.

The failure patterns can also be modelled with statistical distributions. B. Schroeder et al.[4], provided a quantitative analysis of hard disk replacement rates and discussed the statistical properties of the distribution capturing time between replacements. On the other hand, Wang et al. [5], conclude that the time between failure cannot be captured through any of the standard distributions and commented on the difficulty of capturing fault trend using standard distributions.

In [6] deep neural network based RUL (Remaining Useful Life) estimation has been carried out by both LSTM and CNN where LSTM achieved slight better performance. Total 15 S.M.A.R.T metrics are used. The authors chose to perform the tests on the same Seagate model ST4000DM000 and performed the prediction of RUL on models of a single disk with serial number Z300ZQST where the RMSE and the deviation of predicted result from ground truth are shown. The ground truth assumes that the remaining useful time linearly decreases as time elapses. This assumption is not supported by our study shown in Fig.1.

In [7] the trained network using LSTM disk on model ST4000DM000 showed F1-score of 0.77 for 7 days in advance prediction. Xu [8] and Lima [9] also apply LSTM to predict the level of the disk health. Lima et al achieve excellent performance (0.98 of F1 score) for 30 days time in advance. Both use single disk brand in their study.

Existing works focused on selection of features and algorithms. Various types of algorithms have been evaluated. The achieved excellent performance [2, 8, 9] is for single disk brand. In our study, single predictive model is built for disks covering all brands and models. The work of Aussel [3] is in the similar way with data covering multiple brands. Different from previous works, feature selection and algorithm optimization are not our focus. Instead, the basic algorithm and features are fixed in our study to explore fundamental way to improve the performance using derived smart features. The F1 score is 0.94 for a time in advance of 10 days without manual feature engineering, feature selection and fine tuning of the model hyperparameters.

## 3. DATASET

The computer hard driver data from the meter of the Self-Monitoring Analysis and Reporting Technology (S.M.A.R.T or SMART) [10, 11] is collected. It is a monitoring system included in computer hard disk drives (HDD), solid-state drives (SSD), and eMMC drives. The detail of the attributes of S.M.A.R.T data can be found in [12].

We use HDD dataset covering 32 months data from January 2017 to September 2019 [11]. This dataset contains basic hard drive information and 90 columns or raw and normalized values of 45 different SMART statistics. Each row represents a daily snapshot of one hard drive. There is a total of about 86 Million records for about 155k hard disks. Among them, there are 4503 failed disks, which is about 3% of the total disks. Besides SMART statistics, each record has date and tags data such as *serial number, model number, capacity, failure.* This is a typical data structure for IoT (Internet of Things) sensor data: date, tags for device (name, type, location and others) and values. The *serial number* is unique for each driver. Table 1 lists the SMART metrics used in this study. It includes all available features without large missing in the dataset. As we mentioned before, our goal is to build a model with minimized manual feature engineering work.

The raw data is stored in the order of date. Therefore, the history event for each disk should be extracted for each *serial number* and then sorted by date. Null values were replaced with zeros and normalization is done for each SMART attribute. After such data processing, we get event record data for each disk. We call it *sensor event data* in this paper. It is the unit of data set used in the study.

The *model number* is another import tag. There are total 84 models covering different brands. Ideally, disks with the same *model number* can be grouped together to build a model for that particular type of device. However, the dataset for single disk model is typically small and it is not sufficient to build a good analysis model. Instead of building an analysis model for each disk model, we build a robust analysis model for all types of disks because our analysis model can differentiate the normal and failed disk in a deep way to catch the general pattern of all types of disk models. This is one great advantage to leverage the available dataset when they are limited at early stage of the operation.

An important detail is the number days of the lifetime of the replaced. Fig. 1 shows the number days before the disk failed. There is a burst of failure in the first few days when the disks are turned on. This type of failure is related to product defeat, called turn-on failure, that causes the driver to quickly and unpredictably fail. This type of failure is excluded from our prediction model. The second class of failures is the gradual decay of electrical and/or mechanical components within the disk drive. The chance of failure increases with lifetime due to this mechanism. However, it can be seen from the plot that the probability of failure afterward is close to uniform, which indicates a random failure. Furthermore, the lifetime of normal disks due to normal wear is much longer. This import message suggests that the observed disk failure here is not strongly related to normal wear. Therefore, it is not feasible to study the useful lifetime due to normal wear for this type of dataset.

Table 1 Names and IDs of S.M.A.R.T Attributes

| S.M.A.R.T ID | Attribute Name |
|---|---|
| 1 | Read Error Rate |
| 4 | Start/Stop Count |
| 5 | Reallocated Sectors Count |
| 7 | Seek Error Rate |
| 9 | Power-On hours |
| 10 | Spin Retry Count (only for HDD) |
| 12 | Power Cycle Count |
| 183 | SATA Downshift Error Count or Runtime Bad Block |
| 184 | End to End error IOEDC |
| 187 | Reported Uncorrectable Errors |
| 188 | Command Timeout |
| 189 | High Fly Writes |
| 190 | Temperature Difference or Airflow Temperature |
| 191 | G-sense Error Rate |
| 192 | Emergency Retract Cycle Count or Unsafe Shutdown Count |
| 193 | Load Cycle Count or Load Unload Cycle Count Fu... |
| 194 | Temperature or Temperature Celsius |
| 197 | Current Pending Sector Count |
| 198 | Offline Uncorrectable Sector Count |
| 199 | UltraDMA CRC Error Count |
| 240 | Head Flying Hours or Transfer Error Rate Fujitsu |
| 241 | Total LBAs Written |
| 242 | Total LBAs Read |

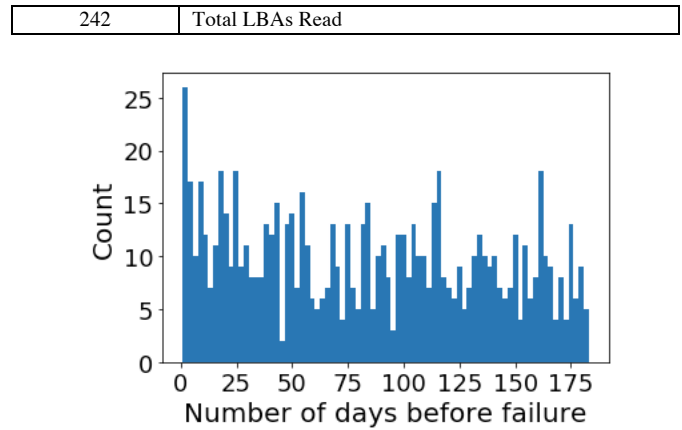

Fig. 1. Number days before the disk failure

## 4. GENERALIZED LEARNING ARCHITECTURE

This section describes the detail neural network module for multivariate time series data with the assumption only multivariate time series data available for the problem like HDD and SDD.

There are many choices in algorithms for failure detection as other problems. For our purpose, the main criteria are automatic feature extraction and capability of handling of various types of data for automation and generalization.

The traditional manual feature engineering build features using domain knowledge. It's a tedious, time-consuming, and error-prone process known. The code with manual feature engineering is problem-dependent and must be re-written for each new dataset.

Automated feature engineering is more efficient and repeatable than manual feature engineering, allowing you to build better predictive models faster. Deep learning-based approaches have been gaining increasing attention in recent years as their parameters can be learned end-to-end without the need for hand-engineered features.

The majority of data set for device and system is continuous number such as the sensor data of device and the reading in Industry Internet of Things (IIOT). In many cases, there are multiple sensors; therefore, they are multivariate time series data. Those time series data are fed to one-dimensional Convolution Neural Network (CNN) for feature extraction. We choose CNN instead of Recurrence Neural Network (RNN) because CNN offers a number of advantages over RNN: simpler model structure, easier to understand and explain. It's much easier to build and train. One advantage of CNN over traditional machine learning is automated feature extraction. The convolution layers automatically generate import features/patterns that are used as input of downstream layers. Therefore, it doesn't require any manual feature engineering.

The Deep Neural Network (DNN) for multivariate time series data is shown in Fig. 2. We use image, a much intuitive way, to visualize the input dataset. Besides the original data (labelled as original metrics), a number of derived features (labelled as smart feature) are feed to the neural network too. The smart features will be explained in next section. The feature extraction layer is 1D CNN layer, which is followed by the transformation layer. In our case, it's another 1D CNN layer with max pooling layer.

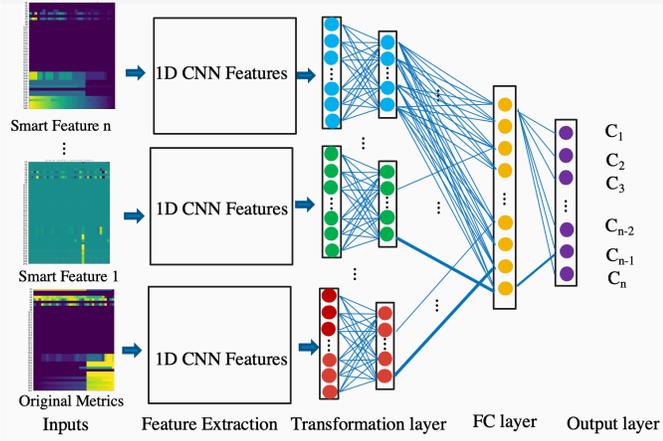

Fig. 2. Architecture of **SFFDD** with smart features of multivariate time series data. The multiple time series data represented by an image with time along horizontal axis.

**Input data as image format**: The data has been preprocessed in such a way that each data record contains *window_length (T)* time slices (each time interval covers one day reading for the case of hard dick drive data). Within each time interval, *Number_features* values (metrics values or sensor readings) are stored. This results in a *window_length × Number_features* (width and height) matrix that is represented by the image in Fig. 2 where the horizontal direction is time.

**First 1D CNN layer (Feature extraction):** The first layer is 1D convolution layer which includes $N_1$ filters (also called feature detector) with a kernel size of $K_1$. It works as a feature extraction layer. Only defining one filter would allow the neural network to learn one single feature in the first layer. This might not be sufficient; therefore, $N_1 \gg 1$ in general. This allows us to train $N_1$ different features on the first layer of the network. The output of the first neural network layer reduces the image width only since the 1D convolution is applied only along the time direction.

**Second 1D CNN layer (1st layer of transformation layer):** The result from the first CNN layer is fed into the second CNN layer which contains $N_2$ filters. The width of image is further reduced.

**Max pooling layer (2nd layer of transformation layer)**: A pooling layer is often used after a CNN layer in order to reduce the complexity of the output and prevent overfitting of the data.

**Fully connected (FC) layer**: After the CNN and pooling, the learned features are flattened to a single vector and is passed through a fully connected layer before the output layer is used to make a prediction. The fully connected layer has $M_{FC}$ neurons.

**Output layer**: The final layer $N_{class}$ neurons where $N_{class}$ is the number of target classes to predict. For binary classification (failure and normal in our case), $N_{class} = 2$. *Softmax* activation function is used for prediction. The *ReLu* activation function is used for the CNN layers and fully connected layer.

In our study we fix the architecture and hyperparameters as the following for comparison purpose: $N_1 = N_2 = 256$, $K_1 = K_2 = 3$, $M_{FC} = 160$, and $N_{class} = 2$. We can also try some more complicated but effective sequence learning models, but in this paper we mainly focus on the idea and effects of smart features, so we just choose a common one for implementation and spend more efforts on explorations of smart features to explore general method of feature derivation for the performance improvement beyond the model optimization.

A system may have multiple types of dataset: number (for example, sensor data), text data (such as log), video and voice. Different types of dataset can be fed to the neural network in separated modules where the features can be extracted and transformed independently as shown in Fig. 3. The transformed features finally merge together and are fed to the output layer for final classification.

The text data is fed to the bedding layer for feature extraction. Various types of embedding methods are available. *Word2vec* is the prominent word embedding technique. *Golve* is another popular one. The recent *BERT* gains better performance in many tasks. The video data is fed to the two-dimensional CNN for feature extraction as it does in many image reignition tasks.

The noise of voice data is first filtered and then features in frequency domain are extracted. Mel Frequency Cepstral Coefficients (MFCC), Linear Prediction Coefficients (LPC), Linear Prediction Cepstral Coefficients (LPCC), Line Spectral Frequencies (LSF), Discrete Wavelet Transform (DWT) and Perceptual Linear Prediction (PLP) are some of the feature extraction techniques used for extracting relevant information from speech signals. The frequency spectrum of the voice of machines is different from speech, but the methods for feature extraction is similar.

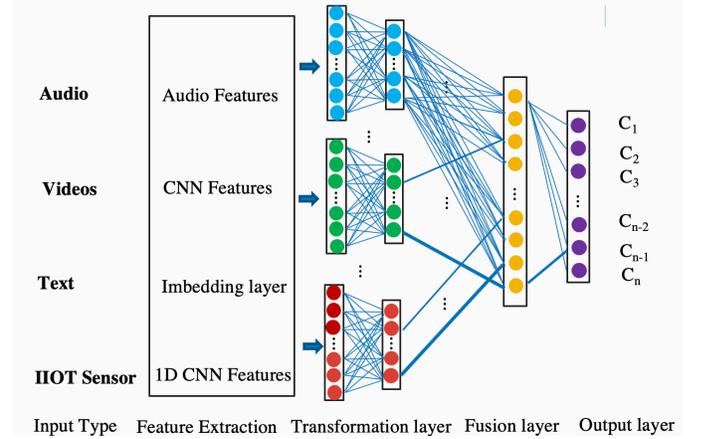

Fig. 3. Architecture of generalized **SFFDD** with multiple types of dataset

## 5. Derivation of New Features

Feature extraction is one of the most import steps in the learning. Instead of doing feature selection, all available S.M.A.R.T metrics are used without manual feature engineering since the feature extraction layer in the deep neural network automatically generates useful features during the learning. We explore novel approaches to deviate new features (called smart features in this paper) especially for failure prediction problem. This approach is inspired by a technique that proved successful in handwritten character recognition (Ha & Bunke [13]). They created extra training data by performing certain operations on real data. In their case, operations like rotation and skew were natural ways to perturb the training data.

The smart features are created by applying various types of pre-defined transformation on the original dataset. Those transformations are specifically chosen to be important for failure prediction problem. The importance of the smart features is evaluated by the model built using them as shown in Fig. 4. It's simple version of Fig. 2. Both the architecture and hyperparameters are fixed without fine tuning for comparison purpose only.

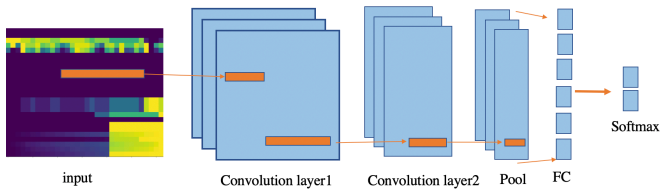

Fig. 4. Architecture of baseline to evaluate the performance of start features. The CNN for multiple time series data represented by an image with time along horizontal axis.

Before we go to the details how to derive smart features, it is intuitive to look at original data as an image. Fig. 5 shows one sample of history record of a failed hard disk. For virtualization purpose, each S.M.A.R.T attribute is normalized into a range between 0 and 255. There are 41 attributes are shown here along the vertical axis with time along horizontal axis. The names with 'normalized' in suffix are special normalization so that higher values indicate healthy disks. Names that end with 'raw' are without such type of normalization. Four patterns in time are clearly shown for different features: oscillated value, gradual change, abrupt change and rare event (with zero value most of the time). The later three patterns are import for failure detection.

### 5.1 Abrupt change by edge detection

Abrupt change or surge is an essential feature in anomaly detection and failure prediction. The change of sensor reading in time or a sudden burst of particular event is an important indicator for failure detection in general. When the reading suddenly changes, it usually indicates an abnormal behavior. This is especially true for electric device. The characteristics of abrupt changes is clearly shown in Fig. 5.

This abrupt change can be better presented by the change over time. Mathematically it's the derivative over time. In practice, this abrupt change can be described in two ways: the change of the value with previous time step; or the normalized change by its current value. The latter is chosen when the value has a trend, like stock price. For most devices and systems, the range of sensor reading value normally should stay within constant range. Therefore, the former is proper choice.

The daily change of the S.M.A.R.T reading can be calculated from the event image shown in Fig. 5 by applying a convolution with kernel [-1,1] along the time axis. This is analogy to the edge detection kernel in image processing domain or rolling difference in finance domain. The event data are stored in an array during the computation, convolution is very convenient and efficient way to apply the transform. It's also straightforward and easy to describe and understand.

The daily change is shown in Fig. 6. The figure becomes clean: only oscillated pattern and abrupt change is clearly visible.

The performance with such daily features is shown in Fig. 12-13. The same architecture shown in Fig. 4 is used for original data and all smart features for comparisons in this section. Balanced dataset is used in the training by randomly selecting the same number of normal disks as the failed ones. The model is trained and validated for 50 times to get the statics of the performance. Surprisingly, the performance is improved compared with baseline with original metrics: better F1-score with balanced performance in both recall and precision, also a smaller spread (variance). This indicts that the abrupt change over time is crucial for sudden/random failure. One can also try to apply other kernels, like [-1,0,1] to find the optimal.

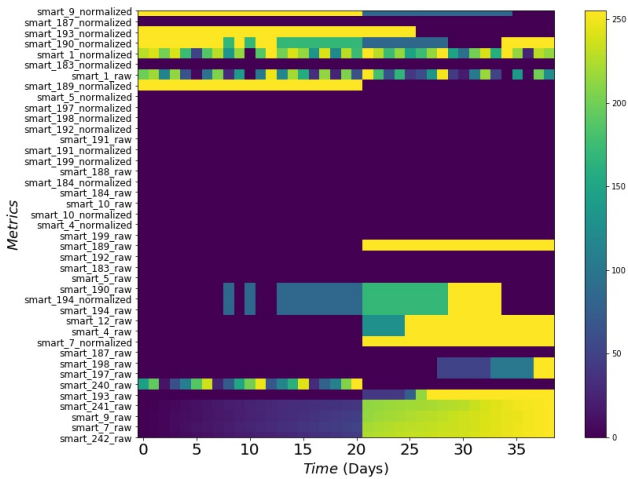

Fig. 5. Visualization of event history of a failed disk with time along horizontal axis and metrics along vertical axis

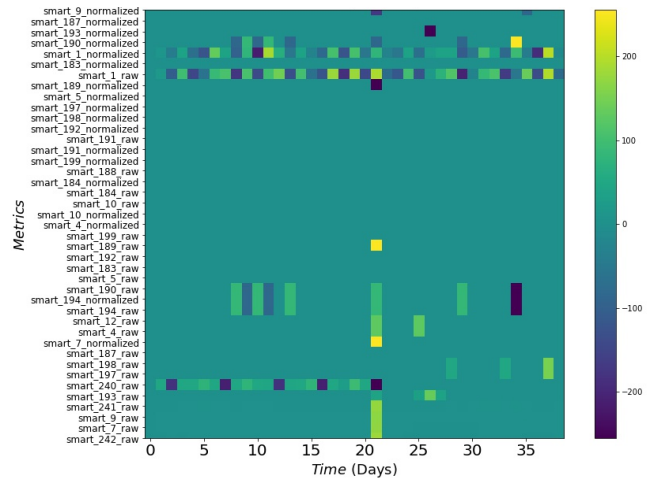

Fig. 6. Visualization of daily change of the failed disk

### 5.2 Smoothing and sharpen kernels

Smoothing can remove the noise from the data. Moving average is one of the popular methods for time series data like finance data to find the long-term trend. Here, we apply a convolution kernel [0.25, 0.5, 0.25] on the event image data in Fig. 5 in the time direction. The smoothed data is shown in Fig. 7. This is similar as the image blurring. The abrupt change is smoothed and becomes slower changes instead of sudden ones. The oscillated features are also smoothed. This type of technique smears out the fast change. Such transformation actually causes partial loss of the useful information in learning which is confirmed by the model. Smoothing helps when longer term trend is important, or the sensor data is too noisy due to measurement precision.

The performance with a burring kernel of [1,4,6,4,1] is slightly degraded comparing with the original features. This indicates abrupt change is a better indicator for this case. A moving average with longer term time window makes a smoother trend line.

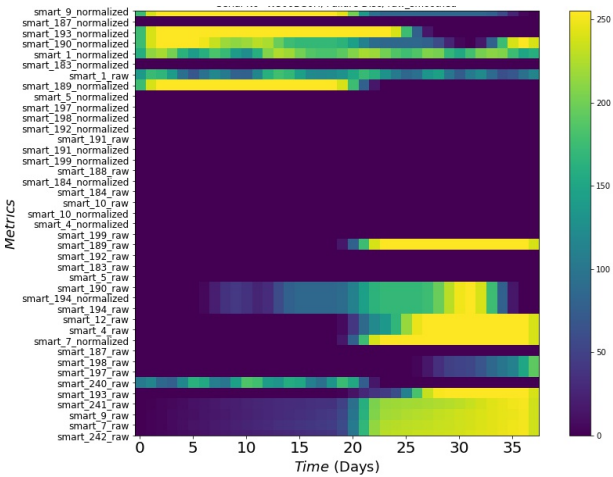

Fig. 7. Visualization of smoothed value by applying a 1D convolution kernel [0.25, 0.5, 0.25]

### 5.3 Accumulative values

Devices can gradually wear out and fail. In this case, the value of cumulative sum in lifetime can be good feature to predict the reminding useful lifetime of a device. Fig. 8 shows the accumulative values. The trend is not clearly visible by this type plot. Instead, the average value of each metrics is clearly shown by the accumulative value at the end date. Each metric of S.M.A.R.T has different physical meaning. Therefore, a relatively large or small value between different metrics is not necessarily important or unimportant.

The performance using accumulative values has the worst performance comparing other smart features in this paper. The accumulative value significantly weakens the importance of abrupt change, which is import for failure learning in general. This may explain the deep insight: the hard driver fails not due to normal wear. This indicator is useful when the failure cause is normal wear where the device usually has a lifetime associated with a normal usage.

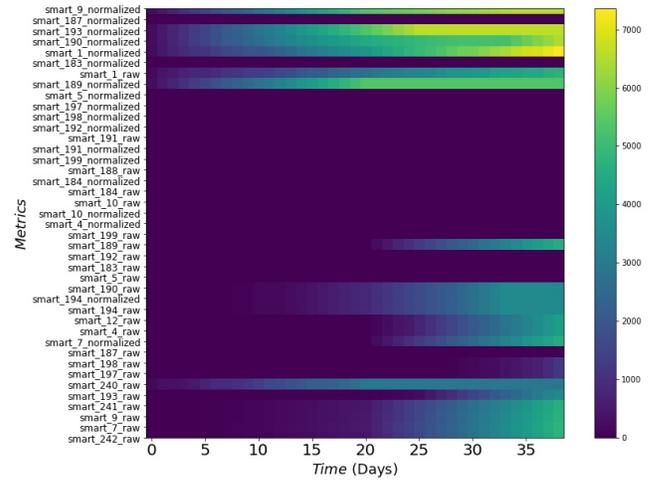

Fig. 8. Visualization of the accumulative values

### 5.4 Reversal arrangement

The overall trend of each metrics can be more clearly seen from the reverse arrangement plot where the counts of reversals, instead of the absolute value, are shown. For particular time, we calculate how many data points of a later arrival time being strictly greater than an earlier arrival time. Each time that happens, we call it a reversal. A large reversal count means a significant up trend.

Fig. 9 shows the reversal arrangement of the disk event. The value of SMART241, SMART9, SMART7 and SMART242 monotonically increases with time. SMART189 behaviors like a step function. It increases suddenly and then keeps constant. SMART1 oscillates but has more chance to decrease after day 5. The non-monotonic trend can be clearly seen. This is one advantage over the plot of original value.

Besides visualization, the reversal arrangement is well used in trend analysis because it represents various trends in a better way. Naturally, it serves as a good measurement for device maintenance and failure analysis since the trend is typically important indicator for the failure or maintenance. Indeed, the analysis model with reversal arrangement has best performance in F1-score among all smart features. It has much higher recall score than that with original features and slightly higher precision than the case with daily changes.

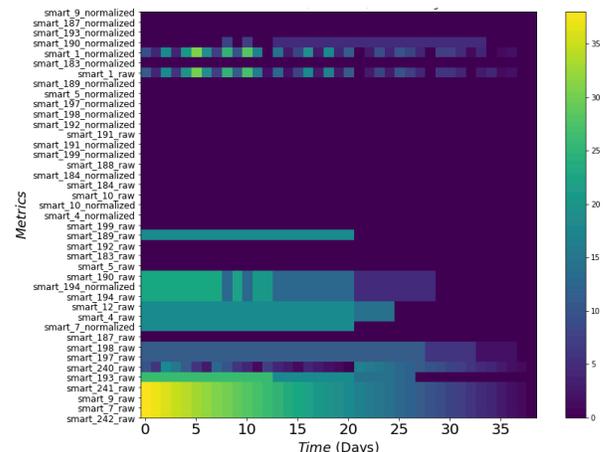

Fig. 9 Visualization of reversal arrangements

## 5.5 CUSUM algorithm

CUSUM algorithm is typically used to detect change in the sample values from a target value (sometimes called mean value). The CUSUM that basically tracks positive ($g+$) as well as negative ($g-$) changes in a temporal sequence ($x$) and checks them against a preset threshold. A change ($alarm$) is detected when $g+$ or $g-$ goes above the threshold. The CUSUM in this scenario can be formulated as follows:

$$g^+(t=0) = 0 \qquad (1)$$
$$g^-(t=0) = 0 \qquad (2)$$
$$s(t) = \mathrm{x}(t) \qquad (3)$$
$$g^+(t) = \max\left(0, g^+(t-1) + s(t) - (Target+K)\right) \qquad (4)$$
$$g^-(t) = \max\left(0, g^-(t-1) - s(t) + (Target-K)\right) \qquad (5)$$

$K$ is usually called the reference value (or the allowance, or the slack value), and is typically set to be equal to half of the distance from the target and the shifted mean [14].

In the traditional application of CUSUM algorithm, the accumulative sum of the changes ($g^{\pm}$) is reset after a change is detected in order to detect next change point. Instead of detecting all changes, we use it to calculate the accumulative sum of change over the whole-time window. The mean value at the initial period is used for Target and $K$ is set to zero. By this way, both the accumulative sum of positive and negative changes for each metrics is calculated as shown in Fig. 10. It can be clearly seen the overall trend comparing with the original data. The metrics which have high values at end of time show strong growth (Fig. 10a) and decline (Fig. 10b). The change points occur at the time when $g^{\pm}$ deviates from zero which indicates abnormal behavior: the values deviate from the target value. The change point is more visible in a line type of plot.

The performance with the accumulative sum of changes ($g^{\pm}$) is slightly better than that using original data, but it's worse than that with edge kernel case. The change location/time is clearly visible by $g^{\pm}$ but the amplitude and the pattern afterwards are not clearly due to the accumulative effect. On the other hand, the result of edge kernel can show both the time/location and amplitude of change at every time step.

Sometimes it is also important to monitor the change in growth. This is called momentum in finance, which indicates the strength of the value change. It can be done simply by using $s(t) = \mathrm{x}(t) - \mathrm{x}(t-1)$ in the above formulae. The results are shown in Fig.11. We name CUSUM sum of change over original value $x$ and daily change $s$ as CUSUM F1 and CUSUM F2, respectively. Various patterns can be identified by combining Fig. 10 and Fig. 11: growth with increasing or decreasing momentum, decline with increasing or decreasing momentum.

Surprisingly, the performance of $g^{\pm}$ over the daily change value is close to the result of daily change. It might be explained by the plot in Fig. 11 where CUSUM F2 better represents the change pattern than CUSUM F1.

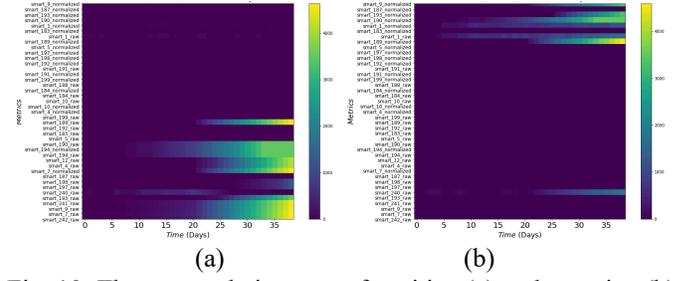

Fig. 10. The accumulative sum of positive (a) and negative (b) changes for original data.

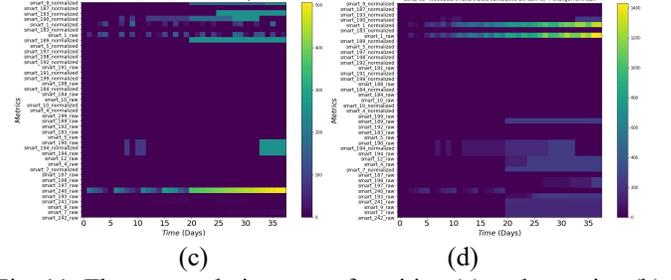

Fig. 11. The accumulative sum of positive (a) and negative (b) changes for daily change.

In short summary, different types of smart features can be generated in such a way by applying pre-defined transformation function on the original data and these new datasets can be used as additional dataset for the learning. We discussed several popular transforms for failure prediction above. There are certainly good transforms available for specific domains or problems. For instance, log-transformation is helpful for exponential trends. The importance of a smart feature for the learning varies with the problems (device or system). It can be easily evaluated by the model fed with them. Table 2 summarizes the patterns monitored by difference smart features.

In the above, a few smart features are discussed and compared in order to better understand their importance and characteristics. The performances with such features are summarized in Fig. 12 and Fig. 13 for the mean and standard derivative value of the performance metrics: F1 score, Recall and Precision.

Reversal arrangement has the best performance with average F1 score of 0.908 and standard derivative of 0.005. Smart feature by edge kernel also works well. CUMSUM F2 also performance well. Furthermore, the three smart features perform well with both high precision and recall. Other smart features have unbalanced performance: higher precision and lower recall. The cumulative value has the worst performance. Original feature has the highest precision but low recall.

Table 2: Patterns represented by Smart Features

| Smart feature | Pattern to monitor |
|---|---|
| Daily Change(Edge kernel) | Abrupt change pattern |
| Moving Average(Blurring kernel) | Noise reduction |
| Cumulative sum | Lifetime related failure |
| Reversal arrangement | Up-trend pattern |
| CUSUM algorithm | Trend up/down from mean |

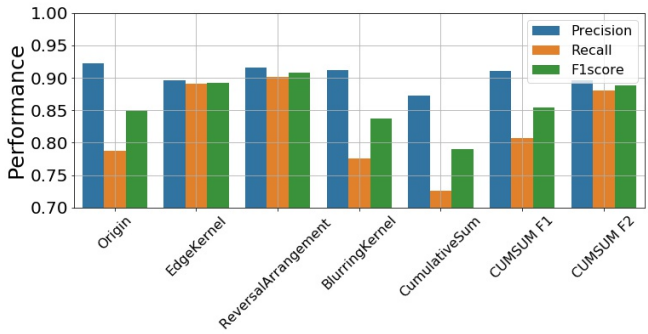

Fig. 12. The mean performance of origin dataset and derived ones. A high score is better.

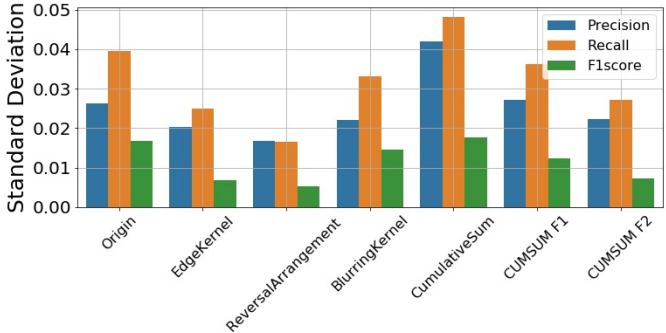

Fig. 13. The standard deviation of the performance with origin dataset and derived ones. A low value is better.

## 6. BUILD A CLASSIFIER WITH MULTIPLE SMART FEATURES

Now we have six smart features. It's ready to build the final model. The architecture for the disk driver is shown in Fig. 2. It's similar as the general architecture proposed in Fig. 3. There are no audio, text and video data in this case. However, we have additional six smart features. For comparison purpose, the same architecture parameters as Fig.4 are used together with all smart features no matter whether they are good indictors for the following reasons. One smart feature alone might not be a perfect indicator; however, it may improve the model if it combines with others. The smart features are derived by different transformation and provide different angle in the learning. Last, the computation time is not a concern, the learning is fast even with all smart features.

The following are the steps to build a single classifier:

1. Prepare data: fetch data from database where the metrics are stored. Then select most recent *window_length* data for each device (unique series number).
2. Processing the data: filling missing value with zero and normalizing each metrics one-by-one. Formatting each device data in matrix format with dimension of *window_length×Number_features* .
3. Smart features generation: apply various transformation on top of event matrix from above step, get smart features.
4. Build a model by feeding original data with smart features together as shown in Fig. 2.

## 7. ENSEMBLE LEARNING

In previous sections, each classifier is built based on the randomly chosen subset of data, called training data. The performance is evaluated using the out-of-sample dataset, called test data. The performance does vary with the training dataset. The variance of the performance indicates the sensitivity to the data sampling. A large variance means large sensitivity. The mean performance is used for comparisons. In practice, which classifier we should use? Should we choose the best classifier? Is there any way to improve the accuracy?

We propose classifier ensemble method to further improve performance. Ensembled classifier with majority vote is an effective way of reducing the variance and improve the accuracy in forecasts. It works as follows: First, $k$ number of classifiers are built based on random sampling of the training and test data with replacement. Second, fit $k$ estimators, one on each training set, with random initial during the training. These estimators are fit independently from each other; hence the models can be fit in parallel. Third, an ensembled classifier that makes a prediction on $N_{class}$ classes by majority vote. The probability that an observation belongs to a class is given by the proportion of estimators that classify that observation as a member of that class (majority voting). This ensembled classifier works as our final model for prediction.

The improvement of ensembled classifier depends on a few factors. First, the potential room to improve decreases when the individual classifiers already perform well (called strong model). For instance, there is no possible improvement in extreme case when individual classifiers work perfectly (with an accuracy of 1). Second, the improvement of ensembled classifier largely depends on the randomness of the classifiers [15]. The more random the better. Fig. 14 shows one example without smart features CUMSUM F1 and F2. There is greater improvement for the ensembled classifier which has the better accuracy than any single classifier. About 100 classifiers are needed to maximize the performance. Fig. 15 compares the accuracy of individual models and the ensembled one. The ensembled model outperforms all individual models in this case. In general, the ensembled model performs better than the average score. The ensembled method offers robust and better performance. There is a similar mechanism for random forest tree algorithm.

When all smart features are used, the final classifier performs degrade slightly due to less gain from the ensemble learning. A strong model in the end doesn't necessarily lead to the best performance with ensemble methods because the gain of ensembled learning depends on the two factors mentioned above. The relationship between different classifiers is not easy to measure and tweak. However, it's definitely an important research topic.

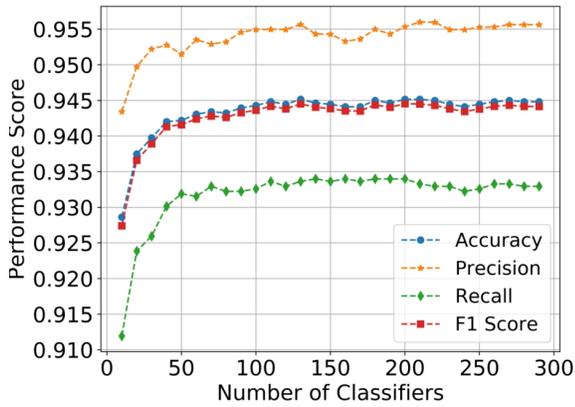

Fig. 14. Performance of ensembled model in the case without smart features CUMSUM F1 and F2

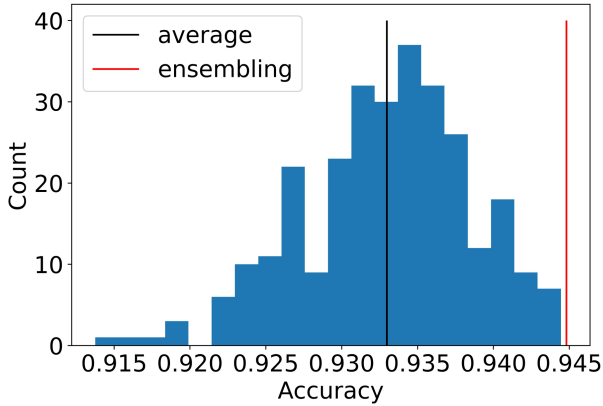

Fig. 15. Performance of individual models and the ensembled model. The accuracy of individual models is shown by the histogram distribution. The mean accuracy is shown by the black line.

## 8. PERFORMANCE OF FAILURE PREDICTION

In previous sessions, an ensembled classifier is built to predict the category of hard disk (failure or normal). The most recent information is used during the training process. For the failed device, the data up to the day before failure is used. This type of classification doesn't have much value since the device fails next day. There is no time to respond to the failure. Therefore, it is important to predict the failure before certain time-window to allow time for maintenance or replacement of the risk device. It is straightforward to predict future failure using the above classifier. The only need is to remove the data of the most recent periods.

To predict $n$ days later whether a disk will fail, the data before $n$ days of the failure is used to build the model. This means that the most recent $n$ days data is removed during the training and test. Then the classifier actually works as a predictor of a failure of $n$ days later.

### 8.1 Prediction of CNN with all smart features

Fig. 16 shows the prediction for different period of time window: 1 day, 10 days and 15 days ahead. The model architecture shown in Fig. 2 is used. Again, all smart features are used to demonstrated automated learning. The performance drops as the time window becomes longer as expected. In practice, there is desired time window time required for maintenance or replacement of the predicted failing device. A time window of 10 days should be sufficient for hard driver replacement. The F1 score is still high as 0.94 for 10 days prediction. Fig. 17 shows the prediction of baseline where only original dataset is used. The model architecture shown in Fig. 4 is used. The F1 score increases from 0.835 by the baseline to 0.94 by SFFDD for 10 days advance. For critical failure prediction, recall is more important than precision. Our SFFDD method largely increases the recall score from 0.78 to 0.927.

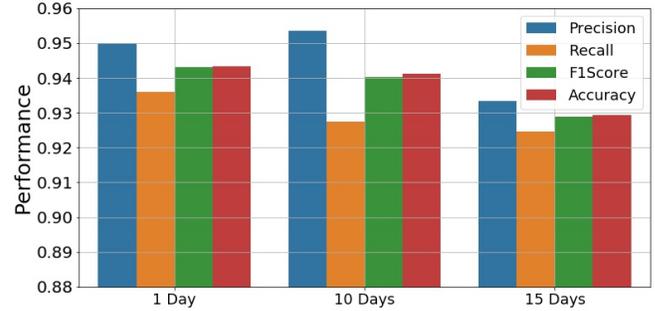

Fig. 16. Failure prediction performance by **SFFDD** with all derived features.

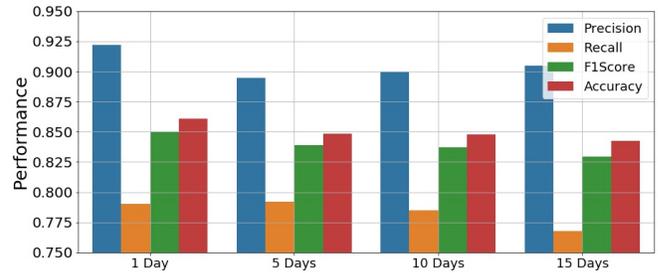

Fig. 17. Failure prediction performance by baseline with only original features

### 8.2 Importance of weak smart features

The smart features are not equally important for the performance as shown in Fig. 12. We suggested to use all them as discussed in previous section. One may optimize smart features to improve the performance. However, it's important to note that it doesn't guarantee better performance by using only good smart features. For illustration purpose, we carry out the following comparison. All smart features are used as shown in Fig.16. In another case, the weak features, including smoothed value, accumulative-sum and CUMSUM F1, are excluded. Its performance (shown in Fig. 18) clearly degrades without those weak features. For instance, F1 score for 15 days prediction drops from 0.929 to 0.90.

We proposed to use all the smart features to leverage weak smart features. Furthermore, we used all available metrics in the dataset shown in Table 1 without feature selection because the feature extraction layer of CNN model automatically extracts import features for us. We found in one test that the model performance doesn't notably degrade by removing some metrics in Table 1. In this case, feature selection can reduce the computation efforts without gain in performance.

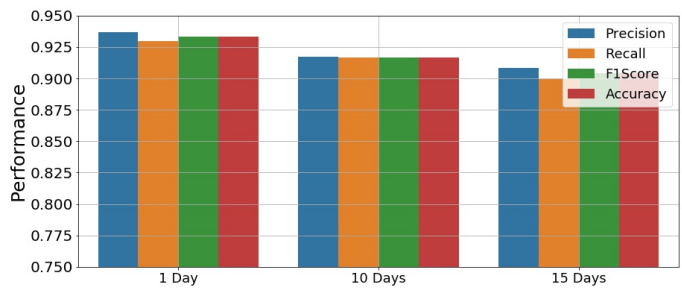

Fig. 18. Prediction performance by **SFFDD** without the weak smart features (smoothed value, accumulative-sum and CUMSUM F1)

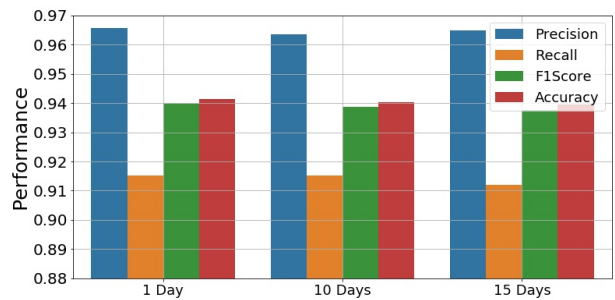

Fig. 20. Performance of LSTM with all derived features. The performance of CNN with the same smart features are shown in Fig. 16.

### 8.3 Comparison with LSTM

LSTM (Long Short-Term Memory) network [16] is a special recurrent neural network. It has long term memory and therefore it performs better for long sequence dynamics. The S.M.A.R.T records, similar as other sensor data, are certainly sequence events. It's nature to use LSTM for such type of data.

Fig. 19 shows the layout of the LSTM model. A bi-directional LSTM with 256 internal units is used. The LSTM block is given an input sequence of $x_1$, $x_2$, ..., $x_T$, of size T, and $x_t$ denotes the entry value at index $t$ in the sequence. In our cases, $x_t$ is the metrics at time $t$ which also includes the derived (SMART) metrics. After the LSTM layer, a pooling layer is used to reduce the array size. Then the learned features are flattened to a single vector and is passed through a fully connected layer before the output layer is used to make a prediction. The same number of FC layers is used as CNN case. The final layer is prediction layer.

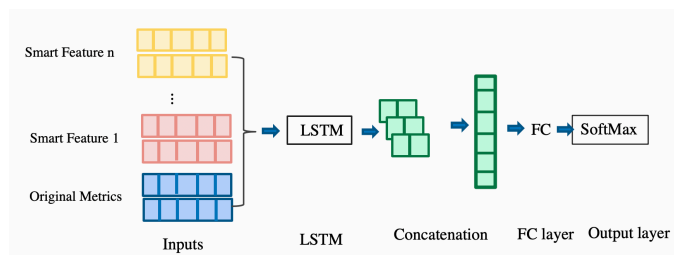

Fig. 19. The framework of LSTM with multiple smart features.

The prediction performance of LSTM is shown in Fig. 19. The performance of LSTM is less balanced between precision and recall. Comparing with the performance of CNN model in Fig. 16, the LSTM has higher score in precision but lower score in recall. For critical failure, recall score in general is more important than precision score in order to detect all failures. CNN performance better in this aspect. On the other hand, LSTM outperforms CNN when precision score is the primary concern. Note that the performance of LSTM degrades very slowly with time in advance. This feature offers great advantage when prediction with longer time in advance is necessary. The performance of LSTM can be superior to that of CNN for longer term prediction. The CNN model is better choice with the assumption that 15 days time in advance is sufficient for hard disk replacement.

### 8.4 Critical time window

It is important to know the essential time window of the information for failure prediction. The abrupt change is an important feature in HDD case. It happens a few days, or more than 10 days before the failure. As a rule of thumb, a time window with double of the that period is the optimal choice: with essential information and less computation effort. The minimum time window should cover the most import pattern, which is the abrupt change for HDD. There is another consideration. The most recent $n$-days data is removed to predict the failure of $n$-days ahead. Therefore, larger time window comparing than $n$ should be used.

It is generally believed that the most recent data contains the most useful information. However, this is much more variable with time series data, in certain time series sequences causality can come from steps much further back (for instance for some rivers it can take 24+ hours for heavy rainfall to raise the river). Significant patterns appear far before HDD fails. The model trained with most recent 7 days dataset performs worse than the model trained using 7 days dataset between 14th to 7th days ago. One explanation is that the abrupt change happens 7 days ago before the failure. Therefore, it is not detectable if only the most recent 7 days data is used. This also explains that the prediction performance drops slowly up to 15 days time in advance. Some patterns (symptoms) appear far before the disk's final failure. In general, the essential time window for failure detection depends on when the critical pattern occurs before the failure.

### 8.5 Model generalization for all brands and models

All results presented in this paper are for all brands and all models of HDD. In other words, the single model works for all HDDs. With our proposed methods, the model with all brands (therefore larger dataset) is always superior than the model trained using single brand dataset even though the variance of the features for all brands is larger. The reasons are twofold: the dataset for single model is relatively small, and our proposed method with multiple smart features can catch the general pattern of all brands and models in a deep way to differentiate the normal and failed device.

Our dataset includes 86 Million records for about 155k hard disks, which covers 84 HDD models. Therefore, the dataset for single HDD model is still relatively small. This is especially true due to the low failure ratio of the device in the dataset. We trained model with the most popular disk model

ST4000DM000 using the same CNN model parameters. The performance is slightly degrades comparing with the model trained with whole dataset. The small dataset is likely the reason. The model with small dataset can be tuned to maximize its performance. For instance, a small model is generally preferred for a small dataset. However, this is beyond the scope of this work.

Model for single HDD device model is trained in the most existing works and the performance is largely sensitive to disk model [2]. There are only a few models trained using all brands of HDD dataset with the best performance of 95% in precision and 67% in recall [3]. Our model has much higher recall of 93% for 15 days prediction.

## 9. Conclusion

A novel feature derivation method specially for device failure prediction is proposed. The method can be applied to devices having sensor type of data, including, but not limited to, computer disk. The smart features are derived by different transformation and provide different angle in the learning. Therefore, the performance of the model is largely improved with these multiple smart features. Furthermore, the model is more robust and generalized covering different brands of the computer disk. The proposed ensemble method further improves the model performance.

The smart features can be directly used by various types of machine learning algorithms. We proposed an architecture of SFFDD based on CNN in order to take various types of dataset together into single analysis model for better prediction. The CNN model outperforms LSTM model in the recall score ( detection of the failed computer disk). On other hand, LSTM has higher precision score and offers an advantage for long term failure prediction. We will explore whether this is true for other devices in the future.